\documentclass[11pt,a4paper]{article}
\usepackage[hyperref]{eacl2021}
\usepackage{times}
\usepackage{latexsym}

\usepackage{multirow}
\usepackage{graphicx}
\usepackage{booktabs}
\usepackage{amsmath}

\usepackage{microtype}

\aclfinalcopy 

\title{SJ\_AJ@DravidianLangTech-EACL2021: Task-Adaptive Pre-Training of Multilingual BERT models for Offensive Language Identification}

\author{Sai Muralidhar Jayanthi \\
    Language Technologies Institute \\
    Carnegie Mellon University \\
  \texttt{sjayanth@cs.cmu.edu} \\\And
  Akshat Gupta \\
  Electrical and Computer Engineering \\
  Carnegie Mellon University \\
  \texttt{akshatgu@andrew.cmu.edu} \\}

\date{}

\begin{document}
\maketitle
\begin{abstract}
In this paper we present our submission for the EACL 2021-Shared Task on Offensive Language Identification in Dravidian languages. Our final system is an ensemble of mBERT and XLM-RoBERTa models which leverage task-adaptive pre-training of multilingual BERT models with a masked language modeling objective. Our system was ranked 1st for Kannada, 2nd for Malayalam and 3rd for Tamil. \footnote{Code and pretrained models are available at  \href{https://github.com/murali1996/eacl2021-OffensEval-Dravidian}{github.com/murali1996/eacl2021-OffensEval-Dravidian}.}
\end{abstract}

\section{Introduction}
\label{sec:introduction}
The task of offensive language identification aims to identify offensive language used in social media texts, which could be in the form of threats, abusive language, insults \cite{zampieri2020semeval}. Social media platforms like Twitter, Facebook, YouTube are now mediums where a large number of people interact and express their opinions. Conflicts and target abuses are common on social media platform. Offensive language detection is a form of social media monitoring which is best done by building computational models and automatic methods due to the large stream of data involved. 

Code-Switching\footnote{Interchangeably used with the term code-mixing} is a phenomenon common in bilingual and multilingual communities where words from two (or more) languages are used in the same sentence \cite{sitaram2019survey} and is very commonly used for social media interactions. Thus it is vital to build models that are able to process code-switched data for the relevant social media tasks. Previous shared tasks in the domain of offensive language identification have focused on identifying offensive language in English \cite{zampieri2019semeval}, Arabic, Greek, Danish and Turkish \cite{zampieri2020semeval}. The datasets used for this shared task in Dravidian languages are characterized by code-mixing.

Pretrained BERT models \cite{devlin2018bert} and their multilingual versions-- mBERT and XLM-RoBERTa \cite{conneau2019unsupervised} have produced state of the art results in multiple tasks. These models have also been used to built best performing sentiment analysis systems in various shared tasks for code-mixed datasets \cite{patwa2020semeval, chakravarthi2020overview}. In this paper we present our submission for the Shared Task on Offensive Language Identification in Dravidian languages. We employ task-adaptive pre-training with a Masked Language Modeling objective and ensemble predictions from mBERT and XLM-RoBERTa models for our final submission.

\section{Dataset}
\label{sec:dataset}
The dataset statistics for the task of Offensive Language Identification in Malayalam \cite{chakravarthi-etal-2020-sentiment}, Tamil \cite{chakravarthi-etal-2020-corpus} and Kannada \cite{hande-etal-2020-kancmd} are shown in Table \ref{tab:stats1}. The class-wise training dataset statistics are shown in Table \ref{tab:stats2}. \citet{hande-etal-2020-kancmd} outlines a brief about the data collection procedure and some baselines. We see that the datasets are highly imbalanced.\footnote{More details of the task can be found at \href{https://competitions.codalab.org/competitions/27654}{https://competitions.codalab.org/competitions/27654}}

\begin{table}[!htbp]
\small
\hspace{0.5cm}
\resizebox{0.4\textwidth}{!}{%
    \begin{tabular}{c|c|c|c}
    \multirow{2}{*}{Language} & \multicolumn{3}{c}{Split} \\ \cline{2-4}
                              & Train   & Dev    & Test   \\ \hline
     Kannada                   & 6217    & 777    & 778   \\
     Tamil                     & 35139   & 4388   & 4392  \\
    Malayalam                 & 16010   & 1999   & 2001   \\
    
    \end{tabular}
}
\caption{Dataset statistics}
\label{tab:stats1}
\end{table}

\begin{table}[!htbp]
\resizebox{0.47\textwidth}{!}{%
    \begin{tabular}{l|c|c|c}
    \multirow{2}{*}{Class} & \multicolumn{3}{c}{Language} \\ \cline{2-4}
                   & Kannada   & Tamil  & Malayalam \\ \hline
    Not-\{language\} &  1522  & 1454   &  1287  \\ \hline
    Not\_offensive  &  3544  & 25425   &  14153  \\ \hline
    Offensive\_Targeted\_Insult\_Individual &  487  &  2343  &  239  \\
    Offensive\_Targeted\_Insult\_Group &  329  &  2557  &  140  \\
    Offensive\_Targeted\_Insult\_Other &  123  &  454  &  0  \\
    Offensive\_Untargetede &  212  &  2906  & 191   \\ \hline
    Total Training Data & 6217   & 35139   &  16010  \\ \hline
    \end{tabular}
}
\caption{Class-wise training data statistics}
\label{tab:stats2}
\end{table}
\section{Modeling}
\label{sec:modeling}

In this section, we describe the different techniques and models that were used in developing offensive language classifiers.
We use two popular transformer architecture based models- Multilingual BERT  \cite{devlin-etal-2019-bert} and XLM-RoBERTa \cite{Conneau_2020} as our backbone models. \cite{gupta2021task} showed that these multilingual BERT models achieve state of the art performance on Dravidian language code-mixed sentiment analysis datasets. Inspired by some of the recent works on Hindi-English codemixed datasets, such as \citet{khanuja-etal-2020-gluecos} and \citet{aguilar-etal-2020-lince}, we develop solutions based on language model pretraining and transliteration.

\subsection{Architectures}

Motivated by the successes of 
BERT model and its underlying Masked Language Modeling (MLM) pretraining strategy~\cite{devlin-etal-2019-bert}, several 
derivatives evolved over the last couple of years. In this task, since the datasets contain texts in English, Tamil, Kannada and Malayalam in their native scripts as well as in Romanized form, we identify two suitable BERT based models for creating offensive identification classifiers. 

Multilingual-BERT\footnote{\href{https://github.com/google-research/bert/blob/master/multilingual.md}{https://github.com/google-research/bert/multilingual.md}}, aka. mBERT, is based on BERT architecture but pretrained with Wikipedias of 104 different languages including English, Tamil, Kannada and Malayalam. In this work, we specifically use the \textit{cased} version of mBERT\footnote{\href{https://huggingface.co/transformers/multilingual.html?highlight=multilingual\#bert}{https://huggingface.co/transformers/multilingual.html}}. mBERT is pretrained on textual data from different languages with a shared architecture and a shared vocabulary, due to which there is a possible representation sharing across native and transliterated text forms. Corroborating it, \citet{pires-etal-2019-multilingual}
presented its superior zero-shot abilities for an NER task
in Hinglish. 

XLM-RoBERTa, on the other hand, is also a transformer based model but is pretrained on Common Crawl data\footnote{\href{http://data.statmt.org/cc-100/}{http://data.statmt.org/cc-100/}}. The common crawl data consists of textual data both in native scripts as well as in romanized script for some of the languages. Among the three languages in this task, only Tamil has Romanized data in the Common Crawl.

Given a stream of text as input, mBERT splits the text into sub-words by using WordPiece tokenization, whereas XLM-RoBERTa uses SentencePiece tokenization strategy. The sub-tokens are then appended with terminal markers such as \texttt{[CLS]} and \texttt{[SEP]}, indicating the start and end of the input stream, as shown below. 
$$ [CLS], w_1, w_2, ..., w_n, [SEP] $$

Popularized by BERT, the representation corresponding to the \texttt{[CLS]} token in the output layer is utilized for multi-class classification via a softmax layer. We utilize the same methodology in this work. While there are alternative choices for obtaining sentence-level representations, such as \textit{sentence-transformers}\footnote{\href{https://github.com/UKPLab/sentence-transformers}{https://github.com/UKPLab/sentence-transformers}}, we leave their evaluation for future work. We obtain results with this setup and consider them as our baselines.

In an attempt to understand the usefulness of character embeddings when augmented with representations from BERT-based models, we develop a fusion architecture. In this architecture, we first obtain sub-word level representations from mBERT or XLM-RoBERTa models. We then convert them back to word-level representations by averaging the representations of sub-tokens of a given word. If a word is not split into sub-tokens, we use its representation as the word's representations. For every batch of training data, we keep track of words and their corresponding sub-tokens for the purpose of averaging. In a parallel, we also pass character-level embedding sequences through a Bidirectional-LSTM for every word. The hope is that the character-level BiLSTM helps to capture different variations in word patterns due to Romanization, especially useful in the context of social media text.

Once we obtain word-level representations from both character-level BiLSTM as well as BERT-based models, we concatenate the representations at word-level and pass them through a word-level Bidirectional-LSTM. The representation of the last word at the output of BiLSTM is then used for classification.

\begin{table*}[!htbp]
\tiny
\centering
\hspace{0.5cm}
\resizebox{0.8\textwidth}{!}{%

\begin{tabular}{cc|ccc}
\multicolumn{1}{c|}{\multirow{2}{*}{Model}} & \multicolumn{1}{c|}{\multirow{2}{*}{Input Text}} & \multicolumn{3}{c}{F1 / Acc (Dev Split)} \\ \cline{3-5}
\multicolumn{1}{c|}{} & \multicolumn{1}{c|}{} & \multicolumn{1}{c|}{Kannada} & \multicolumn{1}{c|}{Tamil} & \multicolumn{1}{c}{Malayalam} \\
\hline

\multicolumn{5}{c}{Baselines} \\ \hline
mBERT & \multicolumn{1}{c}{\multirow{2}{*}{as-is}} & 67.30 / 70.79 & 76.79 / 79.01 & 95.77 / 96.10 \\
XLM-RoBERTa & \multicolumn{1}{c}{} &  68.70 / 71.56 & 76.28 / 78.19 & 94.12 / 94.85  \\
\hline

\multicolumn{5}{c}{Task-adaptive pretraining} \\  \hline
mBERT & \multicolumn{1}{c}{\multirow{2}{*}{as-is}} & 68.40 / 72.07 & 76.95 / 78.87 & 96.58 / 96.75 \\
XLM-RoBERTa & \multicolumn{1}{c}{} & 69.66 / 72.72 & 77.10 / 78.83 & 96.28 / 96.55 \\
\hline

\multicolumn{5}{c}{Transliteration} \\  \hline
mBERT & \multicolumn{1}{c}{\multirow{2}{*}{romanized}} & 68.84 / 72.07 & 75.16 / 78.49 & 95.72 / 95.80 \\
XLM-RoBERTa & \multicolumn{1}{c}{} & 67.73 / 71.17 & 75.73 / 78.08 & 94.18 / 95.00  \\
\hline

\multicolumn{5}{c}{Fusion Architecture (w/ char-BiLSTM)} \\  \hline
XLM-RoBERTa & \multicolumn{1}{c}{\multirow{1}{*}{as-is}} & 69.79 / 72.97 & 75.51 / 77.87 & 95.78 / 96.05 \\
\hline

\multicolumn{5}{c}{Ensemble (Task-adaptive pretraining))} \\  \hline
mBERT & \multicolumn{1}{c}{\multirow{1}{*}{as-is}} & 70.30 / 74.00 & 76.94 / 79.67 & 96.76 / 97.00 \\
\hline

\end{tabular}

}
\caption{Results on all three languages of the task. \textit{as-is} implies the text is inputted without any preprocessing.}
\label{tab:results1}
\end{table*}

\subsection{Training Methodologies}

In this section, we discuss some techniques to tailor mBERT and XLM-ROBERTa models for the task of offensiveness identification in Dravidian languages.

\paragraph{MLM Pretraining:} 

Both the backbone models are pretrained with their respective corpora using Masked Language Modeling (MLM) objective-- given a sentence, the model randomly masks 15\% of the tokens in the input stream, runs the masked stream through several transformer layers, and then has to predict the masked tokens. Doing so helps the model to learn bi-directional contextual representations, unlike LSTM~\cite{doi:10.1162/neco.1997.9.8.1735} based language modeling. 

Recent works such as \citet{Gururangan_2020} showed that a second phase of pretraining (called domain-adaptive pretraining) with in-domain data can lead to performance gains. Relevant in the realm of codemixing NLP, \citet{khanuja-etal-2020-gluecos} and \citet{aguilar-etal-2020-lince} have shown that such a domain-adaptive pretraining can improve performances for classification tasks in Hinglish. They curated a corpora with millions of entries consisting of codemixed Hinglish in Romanized form. Once the multilingual BERT-based models are trained with such data, they utilized those models for downstream tasks. Their downstream tasks generally contained only texts in Romanized form. \citet{Gururangan_2020} have further shown that adapting the models to the task's unlabeled data (called task-adaptive pretraining) improves performance even after domain-adaptive pretraining.

However, we identify few challenges in order to directly adopt ideologies from the related works to our task at hand. Firstly, the datasets provided for each language in this task consists of texts in English, native script of that language and its transliterated form. Thus, in order to convert everything into Romanized form, we need an accurate transliterator tool. Moreover, even if such a oracle transliterator exists, there could be challenges related to word normalization due to peculiarity of a given language's usage in social media platforms. Secondly, obtaining large amounts of Malayalam-English or Kannada-English code-mixed data for the purpose of pretraining is not readily feasible due to scarcity of relevant public datasets. Thus, we resort to only task-adaptive pretraining. 

Some ideas towards curating corpora for domain-adaptive pretraining could be using machine translation datasets which have fair amounts of code-mixing or by adopting semi-supervised code-mixed data creation techniques \cite{gupta-etal-2020-semi}. Alternate approaches could be developing techniques for efficient representation-sharing between words from native script and their Romanized forms, similar to works such as \citet{chaudhary2020dictmlm}. We leave these explorations to future work.

\paragraph{Transliteration:} In order to evaluate the performance of mBERT and XLM-RoBERTa models when inputted with only Romanized script, we first transliterate the task datasets. To this end, we identify \texttt{indic-trans}\footnote{\href{https://github.com/libindic/indic-trans}{https://github.com/libindic/indic-trans}} as the transliterator. While task-adaptive pretraining could also be conducted with transliterated texts, we leave this exploration to future work.

\begin{table*}[!htbp]
\tiny
\centering
\hspace{0.5cm}
\resizebox{0.8\textwidth}{!}{%

\begin{tabular}{l|c|ccc}
\multicolumn{1}{c|}{\multirow{2}{*}{Model}} & \multicolumn{1}{c|}{\multirow{2}{*}{Input Text}} & \multicolumn{3}{c}{F1 / Acc (Dev Split)} \\ \cline{3-5}
\multicolumn{1}{c|}{} & \multicolumn{1}{c|}{} & \multicolumn{1}{c|}{Kannada} & \multicolumn{1}{c|}{Tamil} & \multicolumn{1}{c}{Malayalam} \\
\hline

mBERT (run-1) & \multicolumn{1}{c|}{\multirow{2}{*}{as-is}} & 69.88 / 73.49 & 76.04 / 78.30 & 96.48 / 96.65 \\
XLM-RoBERTa (run-1) & \multicolumn{1}{c|}{} & 69.72 / 72.59 & 76.80 / 79.40 & 95.80 / 96.25\\ 

mBERT (run-2) & \multicolumn{1}{c|}{\multirow{2}{*}{as-is}} &68.67 / 72.72 & 76.58 / 79.17 & 96.50 / 96.70 \\
XLM-RoBERTa (run-2) & \multicolumn{1}{c|}{}  & 68.39 / 71.69 & 76.20 / 79.03 & 96.30 / 96.70\\

mBERT (run-3) & \multicolumn{1}{c|}{\multirow{2}{*}{as-is}} &66.28 / 70.91 & 76.30 / 79.54 & 96.04 / 96.55 \\
XLM-RoBERTa (run-3) & \multicolumn{1}{c|}{} &69.08 / 71.94 & \textbf{77.86} / 78.71 & 96.17 / 96.30 \\

\hline
Ensemble (Mode) & \multicolumn{1}{c|}{\multirow{1}{*}{as-is}} & \textbf{70.30} / \textbf{74.00} & 76.94 / \textbf{79.67} & \textbf{96.76} / \textbf{97.00} \\
\hline

\end{tabular}

}
\caption{Ensemble results on all three languages. \textit{as-is} implies the text is inputted without any preprocessing.}
\label{tab:results2}
\end{table*}
\section{Experiments}
\label{sec:experiments}

\paragraph{Dataset and Metrics:} The number of training and validation examples for each language are presented in Table~\ref{tab:stats1}. Table~\ref{tab:stats2} presents the class-wise distribution of the training data. As observed, the distribution is skewed with the dominant class being \textit{Not\_offensive}. In this work, we experiment with all three languages presented in the shared task. We use F1 and Accuracy as evaluation metrics\footnote{\href{https://scikit-learn.org/stable/modules/generated/sklearn.metrics.classification_report.html}{sklearn.metrics.classification\_report.html}}.

\paragraph{Implementation details:} For the purpose of training and validation, the corresponding data splits are utilized. Since gold labels for test sets were not previously available, different techniques were compared using evaluations on the validation set itself. For the purpose of task-adaptive pretraining using MLM objective, we combine the train and validation splits of the task data for training BERT models, and utilize test split of the task data for testing MLM pretraining. Due to compute resource limitations, we trim longer sentences to 300 characters (including spaces) and always use a batch size of 8. We do not perform any preprocessing on the text so as to allow for a generalized evaluation across the different datasets in the task.

We use Huggingface library~\cite{wolf-etal-2020-transformers} for implementation of our backbone models. Specifically, we use \texttt{bert-base-multilingual-cased} and \texttt{xlm-roberta-base} models. For experiments related to representation fusion, we implement the BiLSTM models in PyTorch~\cite{NEURIPS2019_9015}. During task specific training for offensive classification, we optimize using the 
BertAdam\footnote{\href{https://github.com/cedrickchee/pytorch-pretrained-BERT/blob/master/pytorch_pretrained_bert/optimization.py}{github.com/cedrickchee/pytorch-pretrained-BERT}}
optimizer for models with a BERT component and with Adam \cite{kingma2014adam} optimizer for the remainder. These optimizers are used with default parameter settings. All experiments are conducted for 5 epochs and we pick the best model using the dev set. We use a hidden size of 128 for character-level BiLSTM and a size of 256 for word-level BiLSTM. We add dropouts with 0.40 rate at the outputs of both BiLSTMs.

\paragraph{Ensemble:} For the final submission, we create a \textit{majority voting} ensemble of 6 models-- 3 of mBERT and 3 of XLM-RoBERTa-- based on task-adaptive pretraining technique. Table~\ref{tab:results2} shows the results of various runs and the ensemble scores across all three languages.

\paragraph{Discussion:} We observe that task-adaptive pretraining improves performance (1-2\% absolute) in case of both the BERT-based models, with major gains obtained for Malayalam dataset (upto 2\% absolute improvement). We also observe that the baseline performances are retained when the input text is fully Romanized, thereby opening up scope for future research along the direction of domain-adaptive pretraining. We also observe that the representation fusion helps in some datasets and degrades performance in others. However, we believe that more evaluation needs to be conducted to ascertain the usefulness of such fusion.

\section{Conclusions}
\label{sec:conslusions}
In this paper, we present our submission for EACL-2021 Shared Task on Offensive Language Identification in Dravidian languages. Our system was an ensemble of 6 different mBERT and XLM-RoBERTa models fine-tuned for the task of offensive language identification. The BERT models were pre-trained on given datasets in Malayalam, Tamil and Kannada. Our system ranked 1st for Kannada, 2nd for Malayalam and 3rd for Kannada. We also proposed a fusion-architecture to leverage character-level, subword-level and word-level embedding to improve performance.


\bibliography{anthology,eacl2021}
\bibliographystyle{acl_natbib}


\end{document}